# Fed-NILM: A Federated Learning-based Non-Intrusive Load Monitoring Method for Privacy-Protection

Haijin Wang, *Student Member*, Caomingzhe Si, *Student Member*, Guolong Liu, *Student Member*, Junhua Zhao, *Senior Member*, and Fushuan Wen, *Fellow, IEEE*

*Abstract*--Non-intrusive load monitoring (NILM) is essential for understanding customer's power consumption patterns, and may find wide applications like carbon emission reduction and energy conservation. The training of NILM models requires massive load data containing different types of appliances. However, inadequate load data and the risk of power consumer privacy breaches may be encountered by local data owners during the NILM model training. To prevent such potential risks, a novel NILM method named Fed-NILM which is based on Federated Learning (FL) is proposed in this paper. In Fed-NILM, local model parameters instead of local load data are shared among multiple data owners. The global model is obtained by weighted averaging the parameters. Experiments based on two measured load datasets are conducted to explore the generalization ability of Fed-NILM. Besides, a comparison of Fed-NILM with locally-trained NILMs and the centrally-trained NILM is conducted. The experimental results show that Fed-NILM has superior performance in scalability and convergence. Fed-NILM outperforms locally-trained NILMs operated by local data owners and approximates the centrally-trained NILM which is trained on the entire load dataset without privacy protection. The proposed Fed-NILM significantly improves the co-modelling capabilities of local data owners while protecting power consumers' privacy.

*Index Terms*—Federated Learning, Federated Averaging, Non-intrusive load monitoring, Privacy protection, Sequence-to-point learning.

## I. Introduction

THE total household power consumption can be reduced by 5%-20% through analyzing the appliance-level load information [1]. However, the acquisition of numerous fine-granular appliance-level load data is undoubtedly expensive. Non-intrusive load monitoring (NILM) acquires diverse appliance-level load information through the disaggregation of the total load signal. It is consequently a cost-effective alternative to installing smart meters on each appliance. For most power consumers, currently only the total load is measured by installing advanced metering infrastructures (AMI) at the electrical entry, and the demand for understanding appliance-level behaviors promotes further development of NILM [2]. By acquiring the appliance-level load information, NILM could lay solid foundations for tasks such as energy conservation and emission reduction, etc.

To better implement NILM, a number of methods have been proposed such as expert heuristics algorithms that create rules for each appliance [3], decision trees [4], long-short term memory for event detection [5]. Within the context that the worldwide deployment of AMI brings about abundant load data, research on deep learning-based methods to accomplish NILM is becoming a hot topic in recent years [6-15]. In recent literature, distinct deep neural networks (DNN) structures have been set up to represent the mapping relationship, which includes denoising auto-encoders [7], generative adversarial networks [16], convolutional neural networks (CNN) [9], [12], [17], and recurrent neural networks [9], etc. CNNs have excellent performance among all these methods. However, the selection of hyper-parameters in CNN structures (e.g., number of layers) may affect the ultimate NILM performance. Reference [18] applied background filtering with DNN training to estimate the appliance-level power consumptions. Reference [19] applied deep residual networks for convolutional sequence to sequence learning of NILM, which also improved the performance. Most recently, reference [9] proposed a method based on subtask gated networks that incorporate on-off classification information in addition to the original regression information to outperform the previous best regression result. Among all effective methods applied for NILM, the DNN-based method sequence-to-point (seq2point) learning is commonly used for its outstanding performance in assisting local data owners to accomplish NILM [17].

Deep learning-based methods [7, 8] which have numerous trainable parameters have been developed because plentiful training data yields promising model performance [20]. Besides, to make NILM models generalize to a wider range of power consumers and appliances, it is essential to gather sufficient appliance-level and total power demand data from multiple data owners. Utilities and distribution network operators (DNO) often play the role of local data owners. Collecting massive NILM data is rather costly for local data owners, co-modelling is thus inevitable and beneficial. Though the performance of DNNs can be improved via data cooperation, multi-source load data may cause privacy disputes [21] because there may have chances for a third party to acquire the load data. In particular, load information may be breached when they are saved at a centralized server or in the transmission process. Load data is a core asset for utilities and DNOs, and they are often reluctant to share data due to consumers privacy and assets protection concerns. No existing DNNs have achieved an reasonable trade-off between NILM model performance and user privacy protection [22]. In the smart grid scenario, local differential privacy (LDP) [23] that integrates local obfuscation is widely employed for privacy protection and data control conservation. LDP adds noises selectively on the original load series such that consumers' privacy cannot be easily extracted. Nevertheless, there are vital drawbacks and limitations in LDP. On the one hand, adding excessive noise may distort load data in the co-training process, on the other hand, minor noise may not provide sufficient privacy protection, finding an appropriate amount of noise can be extremely taxing [24]. Therefore, it is tough for LDP to achieve a trade-off between consumer privacy protection and data utilization. Realizing efficient multi-party NILM co-modelling for consumer privacy protection has been increasingly important.

Considering NILM co-modelling among multiple local data owners, it is of great relevance to explore the scalability and convergence of the model. Currently, few works have contributed to this aspect. Without the consideration of the scalability of distributed NILM, more heterogeneous data would bring challenges to the communication burden and the model fitting ability when more local data owners participate in the co-modelling [25]. Reference [26] proposed an effective method dNILM for distributed NILM modelling, but the model performance and convergence have not been thoroughly investigated when more local data owners are involved in co-modelling. With the widespread of AMI and the emergence of DNOs, the scalability and convergence of distributed NILM need to be further examined.

The ever-changing load patterns of industrial customers have introduced new challenges to NILM models. Most current researches are homogeneous, with most literature only examining the NILM performance on appliances in the residential scenario. Few works have conducted NILM analysis in industrial scenarios [6-8], [10]. The validation of NILM model generalization ability in different scenarios is essential to the verification of the model effectiveness.

A privacy-protecting method named Federated Learning (FL) is proposed in [27] to realize co-modelling among several distributed parties. The key idea of FL is to train machine learning models distributionally without training data exchange, so that the privacy of local data can be thus promised. FL has been applied in areas like defense, tele-communications, and pharmaceutics [27].

In this paper, a privacy-protecting method named Fed-NILM is proposed to implement NILM modelling among local data owners, the model performance is guaranteed in the meantime. The principle is to share updated local model parameters between the centralized server and local data owners instead of transmitting load data. Fed-NILM enables local data owners to train models on the overall load dataset, which also reduces the bandwidth requirements in the communication network.

The main contributions of this paper are as follows:

1) This paper establishes an efficient NILM method based on FL and seq2point learning, both model performance and privacy protection are achieved simultaneously. For only model parameters are involved in Fed-NILM, the communication bandwidth required is greatly reduced.

2) Fed-NILM is proved to have superior convergence and scalability. Under the requirement of privacy protection, Fed-NILM performs better when more local data owners are involved.

3) The proposed Fed-NILM has an outstanding performance in generalization. The effectiveness and operability of Fed-NILM are validated in both residential and industrial scenarios.

The remainder of this paper is organized as follows. Section II gives an overall description of Fed-NILM. Section III presents the detailed experiment. Section IV gives the results and analysis. Section V provides the concluding remarks and future works.

## II. ARCHITECTURE AND METHODOLOGIES

This section describes the architecture and methodologies of Fed-NILM. In Fed-NILM, seq2point learning is applied on each local data owner, and Federated Averaging (FedAvg) is employed to accomplish NILM co-modelling. The detailed architecture of Fed-NILM is depicted in Fig. 1.

### A. Objective and Procedure

The objective of Fed-NILM is to protect power consumers' privacy and data control of local data owners in the NILM co-modelling, thus promoting the cooperation among local data owners.

In configured scenarios, a centralized server along with multiple local data owners are set up. It is assumed that there is no power usage information is available on the centralized server, while the global model is generated at the centralized server. Local data owners update their respective model parameters under the coordination of the centralized server. In each round, the centralized server communicates with local data owners to exchange local and global model parameters.

The training and testing of Fed-NILM are outlined as follows,

1) Step 1. A global model is initialized at the centralized server randomly.

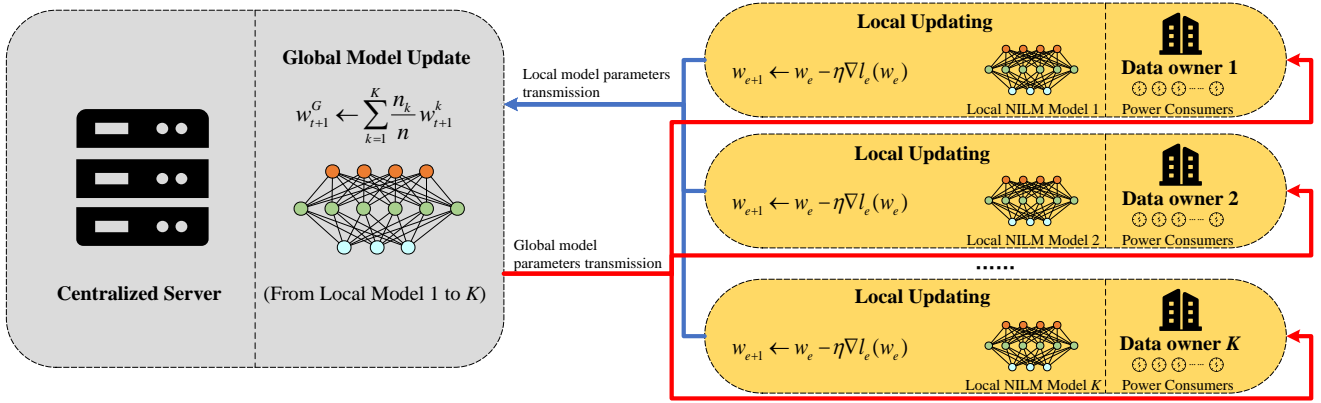

Fig. 1. Detailed architecture of Fed-NILM

2) Step 2. The centralized server transfers global model parameters to local data owners.

3) Step 3. Each local data owner receives the same global model parameters, and the model is trained locally based on each data owner's load dataset individually. After that, each local model parameters are sent back to the server.

4) Step 4. The centralized server weighted averages the local model parameters, then a new global model is trained.

Step 2-4 iterate until a stopping criterion is met, i.e., the maximum number of global training round. The details are introduced in subsection B.

### B. Details of Fed-NILM

*1) Seq2point Learning*

NILM identifies a mapping to represent the functional relationship between the total load signal and the appliance-level ones. With sufficient and quality training load data, DNNs [28] that create networks for the total and appliance-level load signals are particularly suitable for learning the mapping. Among all DNNs, sequence-to-point learning that employs CNNs has superior performance in improving NILM modelling efficiency [10]. Seq2point learning identifies the continuous total load series and decomposes it into multiple appliance-level load series. Seq2point learning is adopted in the "local updating" of Fed-NILM.

Fed-NILM in which combines seq2point learning assumes that the targeted appliances' power usage status at the midpoint element is related to the entirety status before and after the midpoint. In seq2point learning, the input is the total load signal, and the outputs are several appliance-level load signals which are generated at the midpoint of the input temporal window.

NILM can be treated as either a classification task or a regression task [6]. To fully validate the performance of Fed-NILM, the subsequent evaluation is described in terms of classification and regression respectively. In terms of classification, the output signal is the appliance-level power on/off status, and the SoftMax activation function is employed in the dense layer at the bottom of the network structure, while in terms of regression, the output signal is the appliance-level power consumption, and the linear activation function is employed in the dense layer at the bottom of the network structure. The detailed architecture of seq2point learning is illustrated in Fig. 2.

Seq2point learning finds out a network $F_p$ which maps the sliding windows of the total load signals $Y_{s:\,s+W-1}$ to $x_t$. $x_t$ is the midpoint value of windows $X_{s:\,s+W-1}$ for targeted appliance's load signal, $t=s+W/2$.

The mapping $x_t$ is modelled as:

$$x_t = F_p(Y_{s:s+W-1}) + \varepsilon \tag{1}$$

where $\varepsilon$ denotes the $W$ dimensional Gaussian noise.
The loss function for training $L_p$ takes the form:

$$L_p = \sum_{s=1}^{T-W+1} \log p(x_t | Y_{s:s+W-1}, \theta) \tag{2}$$

where $\theta$ are parameters trained by $F_p$. $T$ is the length of total load signals. The advantage of seq2point learning is that there is a single prediction for each $x_t$ other than an average of predictions for each window $Y_{s:\,s+W-1}$.

*2) FedAvg*

Under the privacy protection consideration, Fed-NILM adopts FedAvg to coordinate local data owners that hold differentiated local load datasets for NILM co-modelling without accessing multi-parties' privacy information. The model performance of individual local data owners trained on limited local datasets may not be optimal. FedAvg initializes each local model identically, and each local model is trained locally on their respective local dataset. A weighted average of the local model parameters results in a global model that outperforms the local models. The intuition behind weighted averaging is that local nodes that own a larger dataset have more influence on the shared global model.

There are two essential parts in FedAvg. The first part is "local updating", in which local data owners iteratively update local model parameters. The second part is "global model update", in which the centralized server obtains the global model by weighted averaging the local model parameters. The whole process is run for multiple rounds. Note that during each round of "global model update", the "local updating" process runs for $E$ epochs.

The selected $K$ local data owners iterate "local updating" for $E$ epochs before the "global model update".

At each data owner, the "local updating" is as follows:

$$w_{e+1} \leftarrow w_e - \eta \nabla l_e(w_e) \tag{3}$$

where $\eta$ is the learning rate of "local updating". $w_e$ denotes the local model parameters at epoch $e$, for $e \in [0, E)$. $\nabla l_e$ denotes the obtained gradient at epoch $e$ during "local updating". After $E$ epochs of "local updating", local data owners send updated gradients to the centralized server.

The centralized sever manipulates the "global model update" for $R$ rounds. The centralized server takes a weighted average of the local model parameters, and then transmits the updated global model parameters to local data owners in each round.

At the centralized sever, the "global model update" is as follows:

$$w_{t+1}^G \leftarrow \sum_{k=1}^{K} \frac{n_k}{n} w_{t+1}^k \tag{4}$$

where $n$ and $n_k$ are the number of total training load samples and the number of training load samples on local data owner $k$, respectively. $w_t^k$ are the local model parameters of data owner $k$ at round $t$, $k \in [1, K]$, $t \in [0, R)$. $w_t^G$ are the global model parameters at round $t$.

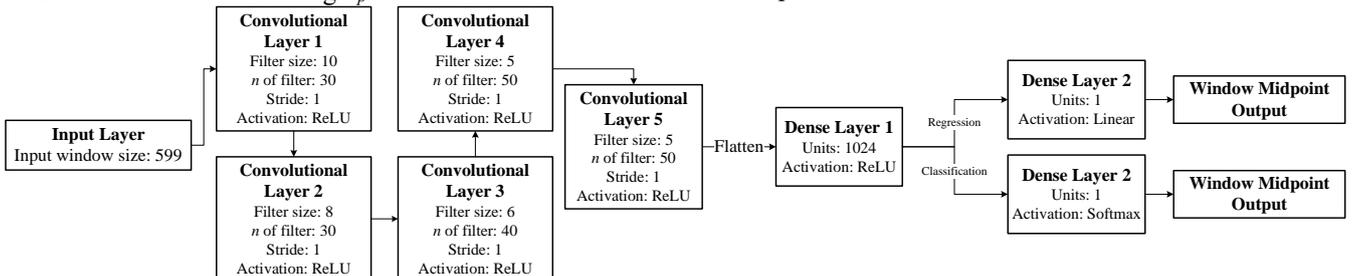

Fig. 2. Seq2point structure in local updating

## C. Performance Evaluation Indicators

Mean absolute error (*MAE*) is a criterion in evaluating Fed-NILM performance. *MAE* reflects how precise the prediction is at every time point. For each targeted appliance, denote $x_t$ and $\hat{x}_t$ as the ground truth and the prediction of power usage status at time $t$, respectively.

The formulation of *MAE* is as follows:

$$MAE = \frac{1}{T}\sum_{t=1}^{T}|\hat{x}_t - x_t| \quad (5)$$

The *MAE* value range in $[0, +\infty)$. *MAE* equals 0 when the prediction exactly matches the true value. The larger the value, the greater the deviation between the prediction and true value.

Another indicator applied in the evaluation is F1 measure which is noted as $F_{1,score}$ hereafter. Given the load series of targeted appliances, the power usage status can be inferred by comparing the consumption with the threshold. $F_{1,score}$ is commonly viewed as the weighted harmonic mean of precision and recall indicators. Precision and recall are mutually constrained in massive load datasets. An ideal performance expects both indicators to be high, but generally, recall is low when precision is high and vice versa.

The math interpretation of $F_{1,score}$ is as follows:

$$F_{1,score} = \frac{2 \cdot TP}{2 \cdot TP + FN + FP} \quad (6)$$

where the value of true positive (*TP*) classifies a positive class as a positive class, the value of true negative (*TN*) classifies a negative class as a negative class, the value of false positive (*FP*) classifies a negative class as a positive class, the value of false negative (*FN*) classifies a positive class as a negative class.

In practical performance evaluation, it is noted as *TP* when the true state of appliance is power-on and is identified as power-on; it is noted as *TN* the true state of appliance is power-on and is identified as power-off; it is noted as *FP* when the true state of appliance is power-off and is identified as power-off; it is noted as *FN* when the true state of appliance is power-off and is identified as power-on.

## D. Model Comparative Indicators

The lower the *MAE* and the higher the $F_{1,score}$, the better the model performs. In terms of both indicators, the experiment presents the improvement of Fed-NILM over the average performance of locally-trained NILMs.

For *MAE* and $F_{1,score}$, the formulations are as follows:

$$\text{Imp}_{MAE} = \frac{Avg(MAE_{loc}) - MAE_{Fed}}{Avg(MAE_{loc})} \quad (7)$$

$$\text{Imp}_F = \frac{Avg(F_{1,score,loc}) - F_{1,score,Fed}}{Avg(F_{1,score,loc})} \quad (8)$$

where $Avg(MAE_{loc})$ is the average *MAE* value of locally-trained NILMs, and $MAE_{Fed}$ is the *MAE* value of Fed-NILM. $\text{Imp}_{MAE}$ represents the improvement of Fed-NILM over the average performance of locally-trained NILMs in terms of *MAE*. $Avg(F_{1,score,loc})$ is the average $F_{1,score}$ value of locally-trained NILMs, and $F_{1,score,Fed}$ is the $F_{1,score}$ value of Fed-NILM. $\text{Imp}_F$ represents the improvement of Fed-NILM over the average performance of locally-trained NILMs in terms of $F_{1,score}$.

The experiment also presents the gap between Fed-NILM and centrally-trained NILM in terms of *MAE* and $F_{1,score}$.

For *MAE* and $F_{1,score}$, the formulations are as follows:

$$\text{Gap}_{MAE} = \frac{MAE_{cent} - MAE_{Fed}}{MAE_{cent}} \quad (9)$$

$$\text{Gap}_F = \frac{F_{1,score,cent} - F_{1,score,Fed}}{F_{1,score,cent}} \quad (10)$$

where $MAE_{cent}$ is the *MAE* of centrally-trained model, and $MAE_{Fed}$ is the *MAE* of the Fed-NILM. $GAP_{MAE}$ represents the gap between Fed-NILM and centrally-trained NILM in terms of *MAE*, $F_{1,score,cent}$ is the $F_{1,score}$ of the centrally-trained model, and $F_{1,score,Fed}$ is the $F_{1,score}$ of the Fed-NILM. $Gap_F$ represents the gap between Fed-NILM and centrally-trained NILM in terms of $F_{1,score}$.

## III. DATASET AND EXPERIMENTS

### A. Data Preparation

In Fed-NILM experiments, the appliance-level and the total load signals are taken from REFIT and IMDELD. The former is a residential load dataset while the latter is an industrial one.

Dataset 1: REFIT is a load dataset measured in Loughborough, England, in which both total and appliance-level consumptions of 20 households from 2013 to 2015 are measured [20]. Due to the huge difference in appliance-level load patterns, REFIT is sufficient for Fed-NILM training in the residential scenario.

Dataset 2: IMDELD is another appliance-level load dataset measured in a poultry feed factory located in Minas Gerais, Brazil from December 2017 to April 2018 [29]. For the factory produces over the year, it has well-behaved usage patterns at any time. IMDELD is used for Fed-NILM training in the industrial scenario.

The residential scenario is designed to explore the convergency and scalability of Fed-NILM. Under the centralized server's coordination, 4, 8, 16, and 32 local data owners are set to accomplish NILM cooperatively. Each data owner holds respective total and appliance-load data. The selected appliances are dishwasher, kettle, microwave, tumble dryer, washing machine, and television. The total and appliance-level load signals are derived from the continuous time-series of building 3, 4, 5, 7, and 19 in the REFIT dataset, ranging from November 1, 2013 to May 1, 2015. Each load signal has 599 dimensions and is sampled every 8 seconds. For each targeted appliance, 162,000 continuous load samples are chosen to build the training set for each local data owner, and 1,296,000 continuous load samples are chosen to build the testing set.

Experiments with different numbers of data owners are carried out to evaluate the generalization ability of Fed-NILM.

The industrial scenario is set for the validation of Fed-NILM. Under the server's coordination, the industrial scenario assumes that 8 local data owners each holds respective total and appliance-level load data. The selected appliances include pelletizer, double-pole contactor, exhaust fan, and milling machine. The total and appliance-level signals are derived from the continuous time-series of building 1 in the IMDELD dataset, ranging from December 11, 2017 to April 1, 2018. Each load signal has 599 dimensions and is sampled every 16 seconds. For each targeted appliances, 5,400 continuous load samples are chosen to build the training set for each local data owner, and 43,200 continuous load samples are chosen to build the testing set.

### B. Experimental Setting

The experiment runs on a server with one 2080Ti GPU and three CPU cores. PyTorch 1.4.0 is used in the experiment.

Seq2point learning [17] is used at each local data owner. A fixed-length window of total load signal is given as input. The input window is generated by sliding the window forward by a single data point. For each targeted appliance, window of the total load signal is the input, while the midpoint status of corresponding windows is the output.

### C. Federated Experiments

This subsection illustrates the conduction of experiment. Both residential and industrial load dataset are chosen to train and test Fed-NILM. The training round in the centralized server is set as 100. For each local data owner, it runs for 10 epochs in each round. ADAM optimizer [39] is applied for model training. Hyper-parameters and optimizer parameters are listed in Table I.

TABLE I HYPER-PARAMETERS AND OPTIMIZER PARAMETERS

| | |
|---|---|
| Input window size | 599 |
| Epoch (for local data owners) | 10 |
| Round (for the centralized server) | 100 |
| Batch size | 512 |

In the early stage of Fed-NILM training, a larger learning rate would make it faster to converge. The learning rate decays with the increase in round, which means that Fed-NILM does not fluctuate significantly, and approximates the convergence steadily. Eventually, Fed-NILM for all appliances are derived. The status of targeted appliance is decided by comparing the consumption with the preset threshold. It is deemed to be power-on if the value exceeds the threshold. The threshold for each appliance is listed in Table II. For different appliances, Fed-NILM is compared with the average level of the locally-trained NILMs and the model trained at the centralized server with the entire training dataset i.e., the centrally-trained NILM.

TABLE II THRESHOLDS FOR APPLIANCES IN THE RESIDENTIAL SCENARIO

| Appliance | Power-on threshold (Watts) |
|---|---|
| Microwave | >200 |
| Washing machine | >5 |
| Kettle | >1400 |
| Dishwasher | >10 |
| Tumble dryer | >130 |
| Television | >10 |
| Pelletizer | >500 |
| Double-pole contactor | >100 |
| Exhaust fan | >1000 |
| Milling machine | >5000 |

## IV. RESULT AND ANALYSIS

### A. Convergence and Scalability Analysis

For the first part, the residential scenario is set up to explore the convergence and scalability of the proposed Fed-NILM. As shown in Fig. 3, the training loss decreases as rounds increases, and the downward trend is independent of the numbers of data owners and appliance types. This demonstrates that the convergence of Fed-NILM is guaranteed in the model training.

For the 6 selected household appliances, the trend in both evaluation indicators of Fed-NILM is depicted in Fig. 4. The black and red curves in Fig. 4 stand for the curve of *MAE* and $F_{1,\text{score}}$ value, respectively, while the coordinates on the left and right side indicate the coordinates of *MAE* and the coordinates of $F_{1,\text{score}}$ indicators, correspondingly. For different appliances and different number of local data owners, Fed-NILM has stable performance. The *MAE* value decreases with the increase of round, while the $F_{1,\text{score}}$ value increases with the increase of round. On the other side, the convergent *MAE* value decreases as the number of data owners increases, and the convergent $F_{1,\text{score}}$ value increases as the number of data owners increases. In Fed-NILM, the more data owners are involved, the better NILM performance it would achieve. Fed-NILM converges more steadily and achieves better performance with more local data owners' participation.

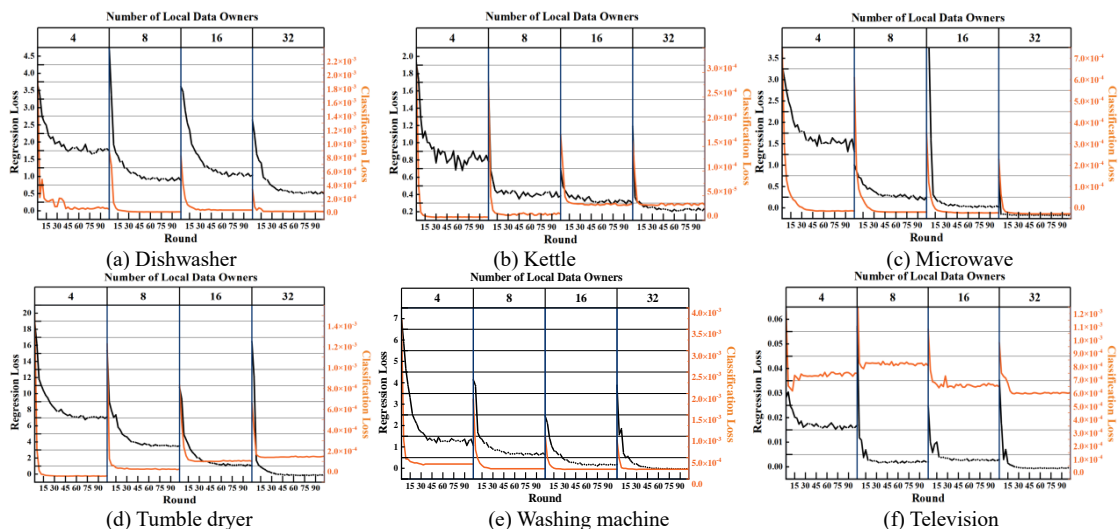

(a) Dishwasher  (b) Kettle  (c) Microwave
(d) Tumble dryer  (e) Washing machine  (f) Television

Fig. 3. Loss vs. Round for 6 household appliances

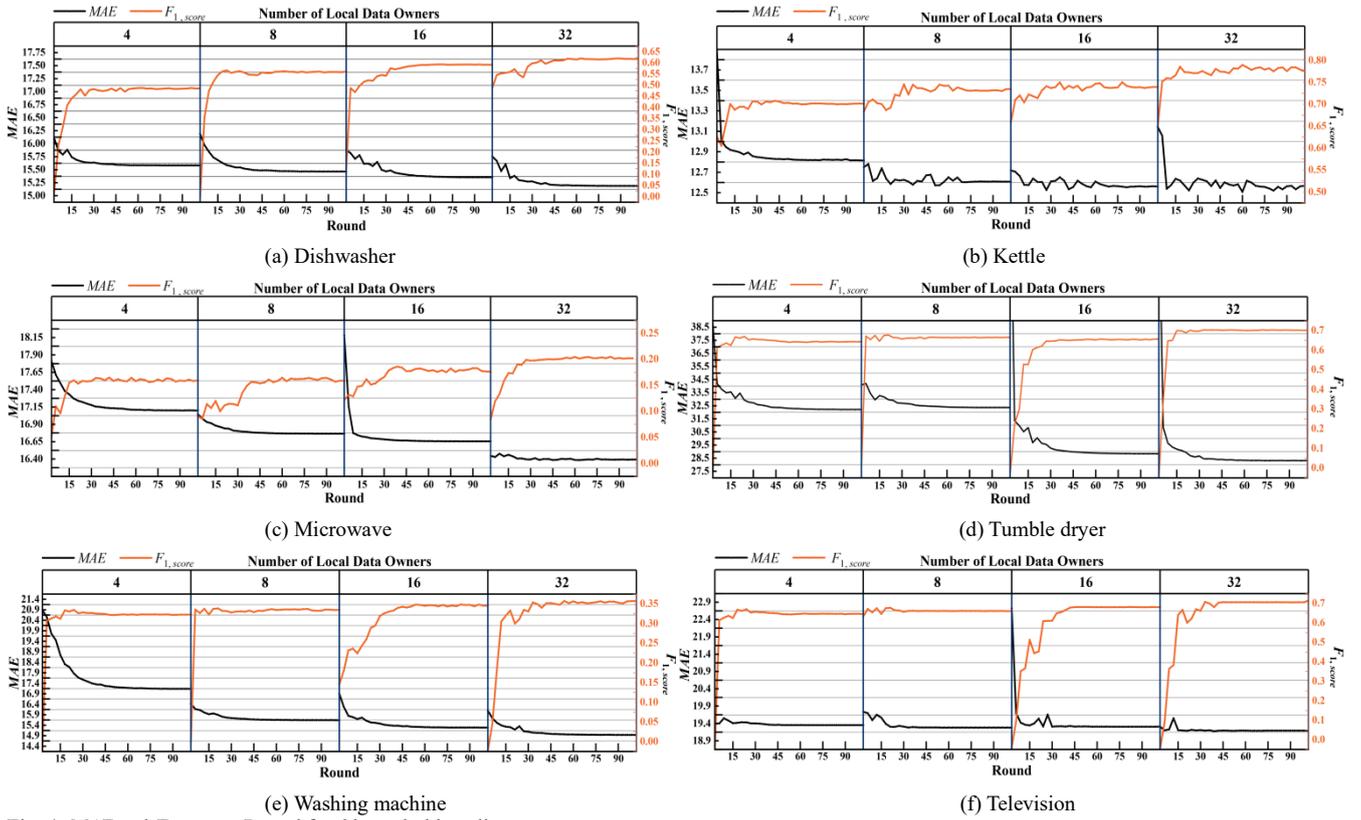

Fig. 4. *MAE* and $F_{1,\text{score}}$ vs. Round for 6 household appliances

TABLE III COMPARISON OF *MAE* INDICATOR FOR APPLIANCES IN DIFFERENT NILM MODELS OF THE RESIDENTIAL SCENARIO

| Appliance | Number of data owners | Locally-trained NILM | | Centrally-trained NILM | | Fed-NILM | | Imp$_{MAE}$ (%) | Imp$_F$ (%) | Gap$_{MAE}$ (%) | Gap$_F$ (%) |
|---|---|---|---|---|---|---|---|---|---|---|---|
| | | MAE | $F_{1,\text{score}}$ | MAE | $F_{1,\text{score}}$ | MAE | $F_{1,\text{score}}$ | | | | |
| Dishwasher | 4 | 56.880 | 0.148 | 15.438 | 0.512 | 15.851 | 0.510 | 72.133 | 244.595 | -2.675 | -0.391 |
| | 8 | 51.213 | 0.160 | 15.087 | 0.588 | 15.734 | 0.586 | 69.277 | 266.250 | -4.288 | -0.340 |
| | 16 | 54.030 | 0.322 | 14.725 | 0.619 | 15.623 | 0.616 | 71.085 | 91.304 | -6.098 | -0.485 |
| | 32 | 50.229 | 0.163 | 14.733 | 0.643 | 15.457 | 0.642 | 69.227 | 295.077 | -4.914 | -0.156 |
| Kettle | 4 | 20.041 | 0.377 | 12.987 | 0.732 | 13.187 | 0.726 | 34.200 | 94.164 | -1.540 | -0.820 |
| | 8 | 20.791 | 0.320 | 12.506 | 0.770 | 12.979 | 0.759 | 37.574 | 140.625 | -3.782 | -1.429 |
| | 16 | 22.471 | 0.290 | 12.157 | 0.774 | 12.938 | 0.764 | 42.424 | 166.897 | -6.424 | -1.292 |
| | 32 | 22.571 | 0.316 | 12.249 | 0.815 | 12.938 | 0.801 | 42.679 | 157.911 | -5.625 | -1.718 |
| Microwave | 4 | 28.065 | 0.049 | 19.725 | 0.189 | 20.360 | 0.185 | 27.454 | 277.551 | -3.219 | -2.116 |
| | 8 | 24.212 | 0.053 | 19.065 | 0.189 | 20.515 | 0.188 | 15.269 | 254.717 | -7.606 | -0.529 |
| | 16 | 23.749 | 0.076 | 16.611 | 0.207 | 16.988 | 0.203 | 28.469 | 167.105 | -2.270 | -1.932 |
| | 32 | 23.920 | 0.081 | 16.458 | 0.230 | 16.664 | 0.226 | 30.334 | 179.012 | -1.252 | -1.739 |
| Tumble dryer | 4 | 108.584 | 0.495 | 32.078 | 0.719 | 32.617 | 0.692 | 69.962 | 39.798 | -1.680 | -3.755 |
| | 8 | 118.074 | 0.480 | 32.023 | 0.727 | 32.772 | 0.714 | 72.245 | 48.750 | -2.339 | -1.788 |
| | 16 | 113.089 | 0.495 | 29.169 | 0.730 | 29.245 | 0.710 | 74.140 | 43.434 | -0.261 | -2.740 |
| | 32 | 115.051 | 0.565 | 27.747 | 0.754 | 28.715 | 0.752 | 75.042 | 33.097 | -3.489 | -0.265 |
| Washing machine | 4 | 39.542 | 0.072 | 17.651 | 0.359 | 17.819 | 0.348 | 54.937 | 383.333 | -0.952 | -3.064 |
| | 8 | 43.527 | 0.115 | 15.479 | 0.364 | 16.325 | 0.359 | 62.495 | 212.174 | -5.465 | -1.374 |
| | 16 | 43.939 | 0.131 | 15.273 | 0.373 | 15.972 | 0.368 | 63.650 | 180.916 | -4.577 | -1.340 |
| | 32 | 38.640 | 0.136 | 15.449 | 0.381 | 15.621 | 0.376 | 59.573 | 176.471 | -1.113 | -1.312 |
| Television | 4 | 20.343 | 0.416 | 18.205 | 0.696 | 18.249 | 0.693 | 10.293 | 66.587 | -0.242 | -0.431 |
| | 8 | 23.606 | 0.360 | 18.132 | 0.727 | 18.462 | 0.710 | 21.791 | 101.944 | -1.820 | -2.338 |
| | 16 | 23.416 | 0.348 | 18.157 | 0.730 | 18.411 | 0.728 | 21.374 | 109.770 | -1.399 | -0.274 |
| | 32 | 22.582 | 0.347 | 17.668 | 0.764 | 18.060 | 0.755 | 20.025 | 117.579 | -2.219 | -1.178 |

TABLE IV COMPARISON OF *MAE* FOR APPLIANCES IN DIFFERENT MODELS OF THE INDUSTRIAL SCENARIO

| Appliances | Locally-trained NILM | | Centrally-trained NILM | | Fed-NILM | | Imp$_{MAE}$ (%) | Imp$_F$ (%) | Gap$_{MAE}$ (%) | Gap$_F$ (%) |
|---|---|---|---|---|---|---|---|---|---|---|
| | MAE | $F_{1,\text{score}}$ | MAE | $F_{1,\text{score}}$ | MAE | $F_{1,\text{score}}$ | | | | |
| Pelletizer | 327.417 | 0.251 | 84.845 | 0.756 | 85.200 | 0.755 | 73.978 | 200.797 | -0.108 | -0.001 |
| Double-pole contactor | 357.421 | 0.254 | 86.230 | 0.754 | 85.124 | 0.751 | 76.184 | 195.669 | 0.309 | -0.004 |
| Exhaust fan | 356.639 | 0.446 | 92.051 | 0.922 | 98.296 | 0.906 | 72.438 | 103.139 | -1.751 | -0.017 |
| Milling machine | 355.726 | 0.462 | 77.980 | 0.925 | 89.901 | 0.910 | 74.727 | 96.970 | -3.351 | -0.016 |

As shown in Tables III, a comparison among Fed-NILM, centrally-trained NILM, and the average level of locally-trained NILMs is made to further illustrate the scalability of Fed-NILM. For *MAE*, centrally-trained NILM often yields lower values than the locally-trained ones; while for $F_{1,\text{score}}$, centrally-trained NILM often yields higher values than the locally-trained ones. In the NILM co-operative modelling process, the limited load dataset scale available to each data owner makes it tough to achieve satisfactory performance of locally-trained NILMs. Fed-NILM can make full use of the entire load dataset while protecting the privacy of each local data owner such that the performance of Fed-NILM is much better than that of locally-trained NILMs.

The model performance of Fed-NILM in both indicators is remarkably approximate to that of centrally-trained NILM, however, Fed-NILM provides significantly better privacy protection for co-modelling participants than centrally-trained NILM. In detail, Fed-NILM achieves a lower value in *MAE* than locally-trained ones and approximates the centrally-trained ones; while the $F_{1,\text{score}}$ value of the global model is higher than that of locally-trained NILMs and approximates the centrally-trained ones.

In contrast to locally-trained NILMs, Fed-NILM has a minimum percentage of improvement in *MAE* at 10.293% and a maximum percentage of improvement at 72.245%, while Fed-NILM has a minimum percentage of improvement in $F_{1,\text{score}}$ at 33.097% and a maximum percentage of improvement at 383.333%. The improvement rate of Fed-NILM over locally-trained NILMs is influenced partly by local load datasets. For exemplified by microwave and tumble dryer, both of which are infrequently used in households, the power-on portion in the load dataset is low, which may in turn severely influences the performance of locally-trained NILMs. Fed-NILM enhances the portion of power-on status relatively, thus reducing the negative impact of imbalanced load datasets. Fed-NILM is superior to the locally-trained NILMs. This can be attributed to that, with the increase of total data in Fed-NILM, the power-on status of appliances with a lower frequency of use increases significantly, which greatly improves the performance of the model.

*B. Generalization Analysis*

Due to differences in the intention of appliance usage in the industrial scenario and the residential scenario, the load characteristics of each appliance vary significantly. To explore the generalization ability of Fed-NILM on the vastly differentiated load datasets, the industrial load dataset, which differs significantly from the residential load dataset, is taken for the experiments. the results of the industrial scenario are presented in Table IV.

The performance of Fed-NILM and centrally-trained NILM is relatively consistent for both indicators, as shown in Table IV. The experimental results demonstrate that Fed-NILM is highly applicable in the context of industrial scenarios. Since industrial appliances operate at much higher power than residential appliances, *MAE* values for all three NILM models obtained in the industrial scenario are generally larger than those in the residential scenario.

The indicator values for each appliance performed varyingly in different scenarios. However, Fed-NILM achieves a significant improvement on the locally-trained NILM and a substantial extent of convergence with respect to the centrally-trained NILM. To summarize, the results and analysis demonstrate the effectiveness and generalization ability of Fed-NILM on varied scenarios and appliances.

The result of the industrial scenario is presented in Table IV. For the four industrial appliances, the improvement in *MAE* and $F_{1,\text{score}}$ of Fed-NILM compared to the locally-trained NILM is significant and approximate the centrally-trained model. The performance of Fed-NILM is comparable to the centrally-trained NILM. This verifies the applicability of Fed-NILM to realistic industrial scenarios.

During the NILM model training process, load data is always maintained by the respective data owners. Fed-NILM can help DNOs and utilities to collaborate in modelling while protecting the consumers' privacy, resulting in a model that approximates the centrally-trained NILM. For Fed-NILM, co-modelling with more data owners can achieve better model performance.

V. CONCLUSIONS AND FUTURE WORKS

This paper establishes a privacy-preserving method named Fed-NILM for multi-party NILM co-modelling. The proposed method eliminates the potential privacy disputes in the co-modelling and facilitates data cooperation among local data owners. For local data owners, seq2point learning is adopted to perform local updating in each round. After acquiring the local model parameters, the centralized server weighted averages the local model parameters to obtain the global model. The ideology of FedAvg is used to assist local data owners in NILM co-modeling.

Both the residential and industrial scenarios are set up in the experiment. The performance of Fed-NILM is evaluated using $F_{1,\text{score}}$ and *MAE*. For both scenarios, one centralized server is responsible for data cooperation among data owners, and distinct appliances are modelled in both scenarios. In Fed-NILM, the risk of consumers' privacy breaches and data control loss is considerably reduced as the local load data are not supposed to leave their respective servers.

The experimental results show that Fed-NILM has a comparable performance compared with the centrally-trained NILM in both residential and industrial scenarios. Moreover, the exceptional performance of Fed-NILM over the locally-trained NILMs has also been proved. In scenarios with different numbers of data owners, it has been found that more local data owners' participation would improve the performance of Fed-NILM, which demonstrates the scalability of Fed-NILM. The proposed Fed-NILM ensures that the global model is stable and convergent under the premise of protecting privacy. To summarize, power consumers' privacy would not be disclosed to other data owners or a third party in the whole NILM co-modelling process while the model performance is maintained relatively.

As for future works, the objective would be to achieve a balance among privacy protection, model performance, and effectiveness. On the premise of privacy protection, finding a more efficient FL method with higher communication efficiency for NILM co-modelling is also a future direction.

REFERENCES

[1] C. Fischer, "Feedback on household electricity consumption: a tool for saving energy?," *Energy efficiency*, vol. 1, no. 1, pp. 79-104, May. 2008.

[2] G. W. Hart, "Non-intrusive appliance load monitoring," in *Proceedings of the IEEE*, vol. 80, no. 12, pp. 1870-1891, Dec. 1992.

[3] M. Weiss, A. Helfenstein, F. Mattern, and T. Staake, "Leveraging smart meter data to recognize home appliances," in *Proceedings of the 2012 IEEE International Conference on Pervasive Computing and Communications*, Lugano, Switzerland, 2012, pp. 190-197.

[4] J. Liao, G. Elafoudi, L. Stankovic and V. Stankovic, "Non-intrusive appliance load monitoring using low-resolution smart meter data," in *2014 IEEE International Conference on Smart Grid Communications (SmartGridComm)*, Aachen, DE, 2014, pp. 535-540.

[5] Q. Liu, K. M. Kamoto, X. Liu, M. Sun, and N. Linge, "Low-complexity non-intrusive load monitoring using unsupervised learning and generalized appliance models," *IEEE Transactions on Consumer Electronics*, vol. 65, no. 1, pp. 28-37, Feb. 2019.

[6] J. Kelly and W. Knottenbelt, "Neural nilm: Deep neural networks applied to energy disaggregation," in *Proceedings of the 2nd ACM International Conference on Embedded Systems for Energy-Efficient Built Environments*, Seoul, South Korea, 2015, pp. 55-64.

[7] R. Bonfigli, A. Felicetti, E. Principi, M. Fagiani, S. Squartini, and F. Piazza, "Denoising autoencoders for non-intrusive load monitoring: improvements and comparative evaluation," *Energy and Buildings*, vol. 158, pp. 1461-1474, Jan. 2018.

[8] J. Kim, T.-T.-H. Le, and H. Kim, "Nonintrusive load monitoring based on advanced deep learning and novel signature," *Computational intelligence and neuroscience*, vol. 2017, pp. 4216281, Oct. 2017.

[9] C. Shin, S. Joo, J. Yim, H. Lee, T. Moon, and W. Rhee, "Subtask gated networks for non-intrusive load monitoring," in *Proceedings of the AAAI Conference on Artificial Intelligence*, Honolulu, USA, 2019, vol. 33, no. 1, pp. 1150-1157.

[10] K. Bao, K. Ibrahimov, M. Wagner, and H. Schmeck, "Enhancing neural non-intrusive load monitoring with generative adversarial networks," *Energy Informatic*, vol. 1, no. 1, pp. 295-302, Oct. 2018.

[11] S. Singh and A. Majumdar, "Deep sparse coding for non-intrusive load monitoring," *IEEE Transactions on Smart Grid*, vol. 9, no. 5, pp. 4669-4678, Sep. 2018.

[12] D. Murray, L. Stankovic, V. Stankovic, S. Lulic, and S. Sladojevic, "Transferability of neural network approaches for low-rate energy disaggregation," in *2019 IEEE International Conference on Acoustics, Speech and Signal Processing (ICASSP)*, Brighton, UK, 2019, pp. 8330-8334.

[13] L. De Baets, J. Ruyssinck, C. Develder, T. Dhaene, and D. Deschrijver, "Appliance classification using VI trajectories and convolutional neural networks," *Energy and Buildings*, vol. 158, pp. 32-36, pp. 32-36, Jan. 2018.

[14] D. García-Pérez, D. Pérez-López, I. Díaz-Blanco, A. González-Muñiz, M. Domínguez-González and A. A. Cuadrado Vega, "Fully-convolutional denoising auto-Encoders for NILM in large non-residential buildings," *IEEE Transactions on Smart Grid*, vol. 12, no. 3, pp. 2722-2731, May 2021.

[15] A. Harell, S. Makonin, and I. V. Bajić, "Wavenilm: a causal neural network for power disaggregation from the complex power signal," in: *2019 IEEE International Conference on Acoustics, Speech and Signal Processing (ICASSP)*, Brighton, UK, 2019, pp. 8335-8339.

[16] Q. Yang, Y. Liu, T. Chen, and Y. Tong, "Federated machine learning: concept and applications," *ACM Transactions on Intelligent Systems and Technology (TIST)*, vol. 10, no. 2, pp. 1-19, Feb. 2019.

[17] C. Zhang, M. Zhong, Z. Wang, N. Goddard, and C. Sutton, "Sequence-to-point learning with neural networks for non-intrusive load monitoring," in *Proceedings of the AAAI Conference on Artificial Intelligence*, New Orleans, USA, 2018, vol. 32, no. 1, pp. 2604-2611.

[18] G. Cui, B. Liu, W. Luan, and Y. Yu, "Estimation of target appliance electricity consumption using background filtering," *IEEE Transactions on Smart Grid*, vol. 10, no. 6, pp. 5920-5929, Nov. 2019.

[19] K. Chen, Q. Wang, Z. He, K. Chen, J. Hu, and J. He, "Convolutional Sequence to Sequence Non-intrusive Load Monitoring," *The Journal of Engineering*, vol. 2018, no. 17, pp. 1860-1864, Nov. 2018.

[20] D. Murray, L. Stankovic, and V. Stankovic, "An electrical load measurements dataset of United Kingdom households from a two-year longitudinal study," *Scientific data*, vol. 4, no. 1, pp. 1-12, Jan. 2017.

[21] S. McLaughlin, P. McDaniel, and W. Aiello, "Protecting consumer privacy from electric load monitoring," in *Proceedings of the 18th ACM Conference on Computer and Communications Security,* ser. *CCS' 11*, New York, USA, 2011, pp. 87-98.

[22] E. Liu and P. Cheng, "Achieving Privacy Protection Using Distributed Load Scheduling: A Randomized Approach," *IEEE Transactions on Smart Grid*, vol. 8, no. 5, pp. 2460-2473, Sept. 2017.

[23] L. Sun, Z. Lin, Y. Xu, F. Wen, C. Zhang, and Y. Xue, "Optimal Skeleton-Network Restoration Considering Generator Start-Up Sequence and Load Pickup," *IEEE Transactions on Smart Grid*, vol. 10, no. 3, pp. 3174-3185, May 2019.

[24] W. S. Choi, M. Tomei, J. R. S. Vicarte, P. K. Hanumolu, and R. Kumar, "Guaranteeing local differential privacy on ultra-low-power systems," in *2018 ACM/IEEE 45th Annual International Symposium on Computer Architecture (ISCA). IEEE*, Los Angeles, USA, 2018, pp. 561-574.

[25] T. Li, A. K. Sahu, A. Talwalkar, and V. Smith, "Federated learning: challenges, methods, and future directions," *IEEE Signal Processing Magazine*, vol. 37, no. 3, pp. 50-60, May 2020.

[26] D. C. Bergman, D. Jin, J. P. Juen, N. Tanaka, C. A. Gunter, and A. K. Wright, "Distributed non-intrusive load monitoring," in: *Innovative Smart Grid Technologies (ISGT)*, Anaheim, USA, 2011, pp. 1-8.

[27] B. McMahan, E. Moore, D. Ramage, S. Hampson, and B. A. y Arcas, "Communication-efficient learning of deep networks from decentralized data," in: *Proceedings of the 20th International Conference on Artificial Intelligence and Statistics, PMLR*, Fort Lauderdale, USA, 2017, vol. 54, pp. 1273-1282.

[28] J. Kim, T.-T.-H. Le, and H. Kim, "Non-intrusive load monitoring based on advanced deep learning and novel signature." *Computational intelligence and neuroscience*, vol. 2017, pp. 4216281, Oct. 2017.

[29] P. B. de Mello Martins, V. B. Nascimento, A. R. de Freitas, P. B. e Silva, and R. G. Duarte Pinto, "Industrial machines dataset for electrical load disaggregation," *IEEE Dataport*, Dec. 2018.